\documentclass{bmvc2k}

\usepackage{amssymb}
\usepackage{multirow}
\usepackage{bbm}

\title{Cascade RetinaNet: \\ Maintaining Consistency for Single-Stage Object Detection}

\addauthor{Hongkai Zhang}{hongkai.zhang@vipl.ict.ac.cn}{1,2}
\addauthor{Hong Chang}{changhong@ict.ac.cn}{1,2}
\addauthor{Bingpeng Ma}{bpma@ucas.ac.cn}{2}
\addauthor{Shiguang Shan}{sgshan@ict.ac.cn}{1,2,3}
\addauthor{Xilin Chen}{xlchen@ict.ac.cn}{1,2}

\addinstitution{
Key Laboratory of Intelligent Information Processing of Chinese Academy of Sciences (CAS),\\
Institute of Computing Technology, CAS,\\
Beijing, 100190, China
}
\addinstitution{
University of Chinese Academy of Sciences,\\
Beijing, 100049, China
}
\addinstitution{
CAS Center for Excellence in Brain Science and Intelligence Technology,\\
Shanghai, 200031, China
}

\runninghead{H. Zhang, H. Chang, B. Ma, S. Shan, X. Chen}{Cascade RetinaNet}

\def\eg{\emph{e.g}\bmvaOneDot}

\def\etal{\emph{et al}\bmvaOneDot}

\begin{document}

\maketitle

\begin{abstract}

Recent researches attempt to improve the detection performance by adopting the idea of cascade for single-stage detectors. In this paper, we analyze and discover that inconsistency is the major factor limiting the performance. The refined anchors are associated with the feature extracted from the previous location and the classifier is confused by misaligned classification and localization. Further, we point out two main designing rules for the cascade manner: \textit{improving consistency between classification confidence and localization performance}, and \textit{maintaining feature consistency between different stages}. A multistage object detector named Cas-RetinaNet, is then proposed for reducing the misalignments. It consists of sequential stages trained with increasing IoU thresholds for improving the correlation, and a novel Feature Consistency Module for mitigating the feature inconsistency. Experiments show that our proposed Cas-RetinaNet achieves stable performance gains across different models and input scales. Specifically, our method improves RetinaNet from 39.1 AP to 41.1 AP on the challenging MS COCO dataset without any bells or whistles.

\end{abstract}

\section{Introduction}

Object detection serves as a fundamental task in computer vision field which has made remarkable progress by deep learning in recent years. Modern detection pipelines can be divided into two major categories of one-stage detection and two-stage detection. Generally speaking, two-stage methods (\eg {Faster R-CNN~\cite{FasterRCNN}}) have been the leading paradigm with top performance. As a comparison, one-stage approaches (\eg {YOLO~\cite{YOLO} and SSD~\cite{SSD}}) which aim at achieving real-time speed while maintaining great performance are attracting more and more attention.

Recent researches focus on improving detection performance from various perspectives~\cite{GIoU, Trident, Regionlets, CornerNet, Revisiting}. A simple idea is adding new stages for additional classifications and regressions which leads to more accurate confidence scores and higher localization performance. Cascade R-CNN~\cite{CascadeRCNN} improves two-stage methods by utilizing cascade sub-networks for gradually increasing the quality of region proposals. As for one-stage methods, RefineDet~\cite{RefineDet} adopts a refinement module to simulate the second regression as in two-stage methods. Consistent Optimization~\cite{ConsistentOptimization} attaches subsequent classification targets for the regressed anchors which reduce the gap between training and testing phases. However, cascade-like single-stage methods ignore the \textit{feature consistency} which limits their effectiveness. For instance, RetinaNet~\cite{FocalLoss}, the state-of-the-art one-stage detection pipeline, generates anchors from feature pyramids and performs classification and regression for each anchor using the feature extracted at the anchor's center point. If we add cascade stages to RetinaNet, the output anchors of the first stage will have shifted center points compared with the original ones. Since most single-stage methods perform feature extraction via sliding window based on the original location instead of the regressed location, feature inconsistency inevitably occurs between different cascade stages. 

In this paper, we discover that naively cascading more stages with the same setting as the original one brings no gains for RetinaNet. The main reasons are two-fold: the mismatched correlation between classification confidence and localization performance, and the feature inconsistency in different stages. In RetinaNet, anchors are regarded as positive if its intersection-over-union (IoU) with a ground-truth is higher than a threshold (\eg{0.5}). It means that no matter the actual IoU is 0.55 or 0.95, the classification targets are the same. So the classification confidence can not reflect the localization performance as mentioned in IoU-Net~\cite{IoUNet}. We find that the mismatched correlation problem can be naturally addressed in a cascade manner by gradually raising the IoU thresholds for the latter stages since the targets are more consistent with the actual IoU. To deal with the feature misalignments, a simple but effective Feature Consistency Module (FCM) is introduced for adapting the features to the refined locations. Specifically, the offset for each location on the feature map is predicted and a simple deformable convolution~\cite{Deformable} layer is utilized to generate the refined feature map for the following stage. In this cascade manner, a sequence of detectors adapted to increasingly higher IoUs can be effectively trained and the detection results can be refined gradually.

The main contributions of this work are summarized as follows:
\begin{itemize}
    \item We revisit the feature inconsistency problem in recent researches and point out two main designing rules for cascade single-stage object detection: \textit{improving the consistency between classification confidence and localization performance}, and \textit{maintaining feature consistency between different stages}.
    \item To improve the reliability of classification confidence, IoU thresholds are increased gradually in the cascade manner. FCM is also introduced to mitigate the feature inconsistency between different stages.
    \item Without any bells or whistles, our proposed Cas-RetinaNet achieves stable performance gains over the state-of-the-art RetinaNet detector.
\end{itemize}

\section{Related Work}
\textbf{Classic object detectors.}
In advance of the wide development of deep convolutional networks, the sliding-window paradigm dominates the field of object detection for years. Most progress is related to handcrafted image descriptors such as HOG~\cite{HOG} and SIFT~\cite{SIFT}. Based on these powerful features, DPMs~\cite{DPM} help to extend dense detectors to more general object categories and achieves top results on PASCAL VOC~\cite{PASCAL}.

\textbf{Two-stage object detectors.}
In the modern era of object detection, Faster R-CNN~\cite{FasterRCNN}, on representative of two-stage approaches, has been the leading paradigm with top performance on various benchmarks~\cite{COCO, PASCAL, Wider2018}. Several extensions to this framework have been proposed to boost the performance, including adopting multi-task learning scheme~\cite{MaskRCNN}, building feature pyramid~\cite{FPN}, and utilizing cascade manner~\cite{CascadeRCNN}.

\textbf{One-stage object detectors.}
Compared with two-stage methods, one-stage approaches aim at achieving real-time speed while maintaining great performance. OverFeat~\cite{OverFeat} is one of the first modern single-stage object detectors based on deep networks. YOLO~\cite{YOLO, YOLO9000} and SSD~\cite{SSD} have renewed interest in one-stage approaches by skipping the region proposal generation step and directly predicting classification scores and bounding box regression offsets. Recently, Lin~\etal point out that the extreme foreground-background class imbalance limits the performance and propose Focal Loss~\cite{FocalLoss} to boost accuracy. Generally speaking, most one-stage detectors follow the sliding window scheme and rely on the fully convolutional networks to predict scores and offsets at each localization which is beneficial to reduce the computational complexity.

\textbf{Misaligned classification and localization accuracy.}
Non-maximum suppression (NMS) has been an essential component for removing duplicated bounding boxes in most object detectors since~\cite{HOG}. It works in an iterative manner. At each iteration, the bounding box with the maximum classification confidence is selected and its neighboring boxes are suppressed using a predefined IoU threshold. As mentioned in~\cite{IoUNet}, the misalignment between classification confidence and localization accuracy may lead to accurately localized bounding boxes being suppressed by less accurate ones in the NMS procedure. So IoU-Net~\cite{IoUNet} predicts IoU scores for the proposals to reduce this misalignment.

\textbf{Cascaded classification and regression.}
Cascading multiple stages is a simple idea to obtain more accurate confidence and higher localization performance. There have been attempts~\cite{CascadeRCNN, ConsistentOptimization, CascadeRPN, RefineDet, FA-RPN, DeepProposal} that apply cascade-like manner to reject easy samples at early stages, and perform bounding box regression iteratively. However, conventional methods (especially the one-stage ones) ignore the feature consistency between different cascade stages since most of them extract features from the original position using a fully convolutional manner. Two-stage detectors generate predictions based on the region features extracted by RoI-Pooling~\cite{FastRCNN} or RoI-Align~\cite{MaskRCNN}. These operations reduce the misalignment between stages since the feature does not correlate with the anchor centers strongly. As for the one-stage approaches, sliding window scheme leads to well alignments between anchor feature and anchor centers. Refined anchors for the next stage are associated with the feature extracted from the previous location, which leads to limited detection performance.

\section{Analysis in Cascade Manner}

\begin{figure}[!t]
\centering
\subfigure[First Stage]{
    \begin{minipage}{0.47\linewidth}
        \centering
        \includegraphics[width=\linewidth]{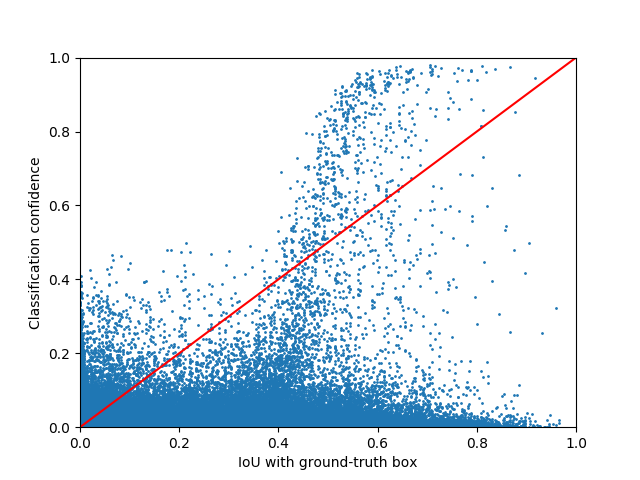}
    \end{minipage}
}
\subfigure[Second Stage]{
    \begin{minipage}{0.47\linewidth}
        \centering
        \includegraphics[width=\linewidth]{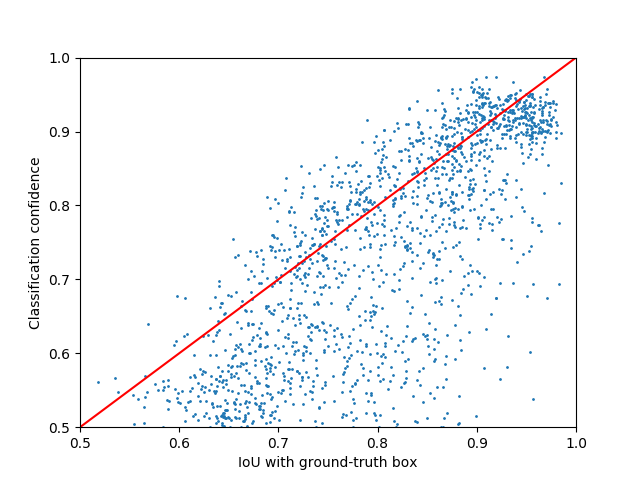}
    \end{minipage}
}
\caption{The correlation between the IoU of bounding boxes with the matched ground-truth and the classification confidence for different cascade stages. The red line represents the ideal situation. (a) Misalignment in the first stage, especially for the confidences near IoU@0.5. (b) Improved consistency between classification and regression in the second stage using increased IoU threshold.}
\label{fig:distribution}
\vspace{-0.3cm}
\end{figure}

In this section, we mainly talk about a simple but vital question: \textit{what kind of stages can be cascaded in single-stage architecture}? From our perspective, there are two pivotal designing rules: improving consistency between classification confidence and localization performance, and maintaining feature consistency between stages.

\subsection{Misaligned Classification and Localization}
\label{sec:misaligned_cls}

Generally speaking, performing classification and regression multiple times can gradually improve the results especially the localization performance for two-stage detectors~\cite{CascadeRCNN}. However, we find that simply adding extra stages with the same setting as the original one does not work for single-stage detectors. During the analysis, we find that the reason for this phenomenon mainly lies in \textit{the inconsistency between classification confidence and localization performance}. In cascade single-stage detector, pre-defined anchors are used as the input of the first stage, and regression offsets are added to generate the refined anchors which are viewed as the input of the second stage. As illustrated in Figure~\ref{fig:distribution} (a), the bounding boxes with higher IoU are not well associated with higher classification confidences in the first stage, especially for the confidences near IoU@$0.5$. 
The misaligned confidences lead to confused ranking which limits the overall performance. 

In order to reduce this negative effect, we change the decision condition of positive samples for the following stages by increasing the IoU thresholds, such that samples with higher quality are chosen as positive. 
However, excessively large IoU thresholds lead to exponentially smaller numbers of positive training samples, which can degrade detection performance~\cite{CascadeRCNN}. From our experiments, we find that gradually increasing the IoU threshold leads to boosted performances.

\subsection{Feature Inconsistency}

Most single-stage methods perform feature extraction via sliding window based on the anchor location. The sliding window schemes obtain well alignments between anchor feature and anchor centers since the features are extracted in a fully convolutional manner. For instance, RetinaNet attaches a small fully convolutional network which consists of four convolutions for feature extraction and a single convolution layer for prediction in different branches. The prediction of each position on the feature map contains classification and regression for various anchor shapes.

\begin{figure}[!t]
\centering
\subfigure[Original image]{
    \begin{minipage}{0.42\linewidth}
        \centering
        \includegraphics[width=\linewidth]{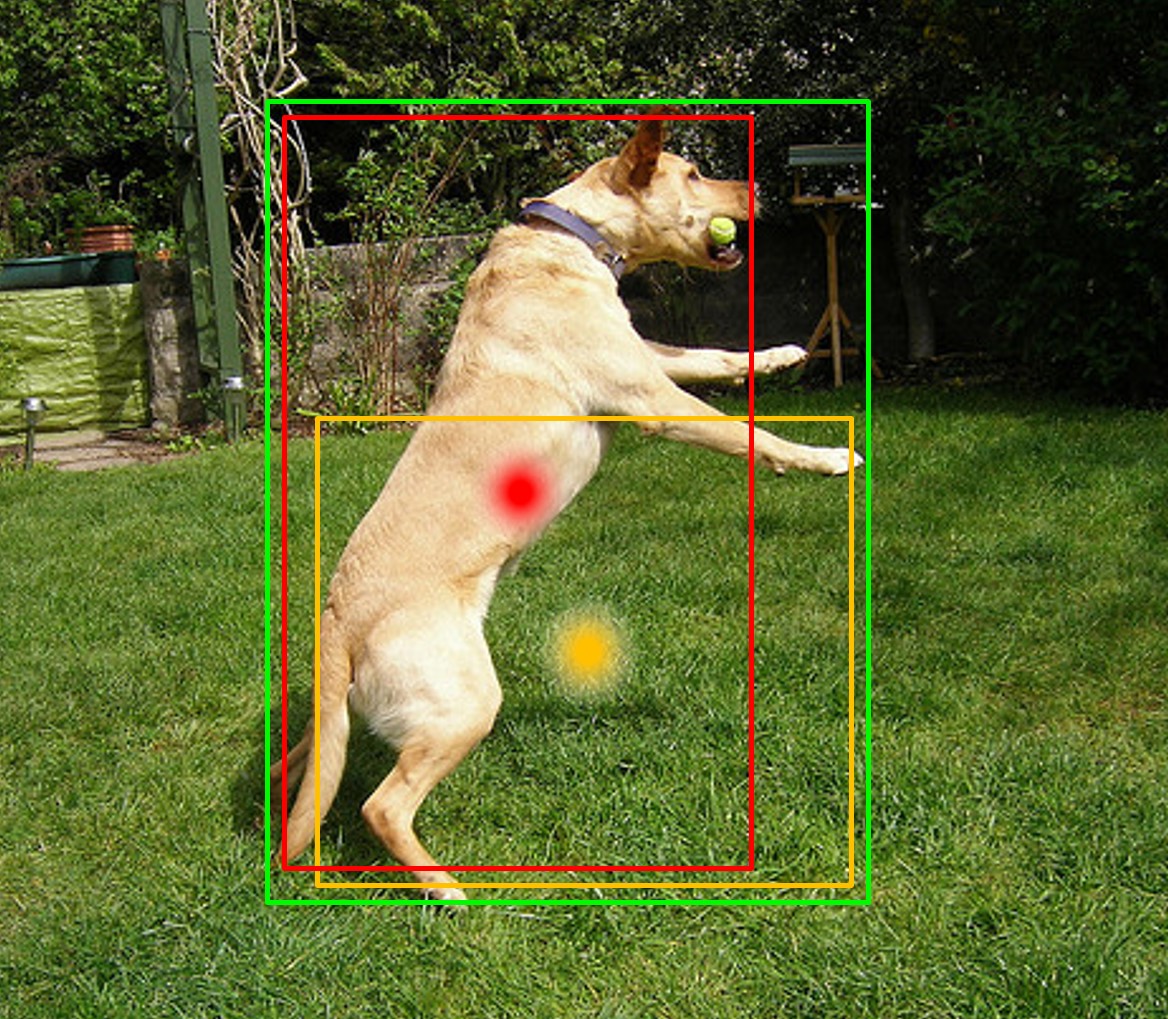}
    \end{minipage}
}
\subfigure[Feature grid]{
    \begin{minipage}{0.42\linewidth}
        \centering
        \includegraphics[width=\linewidth]{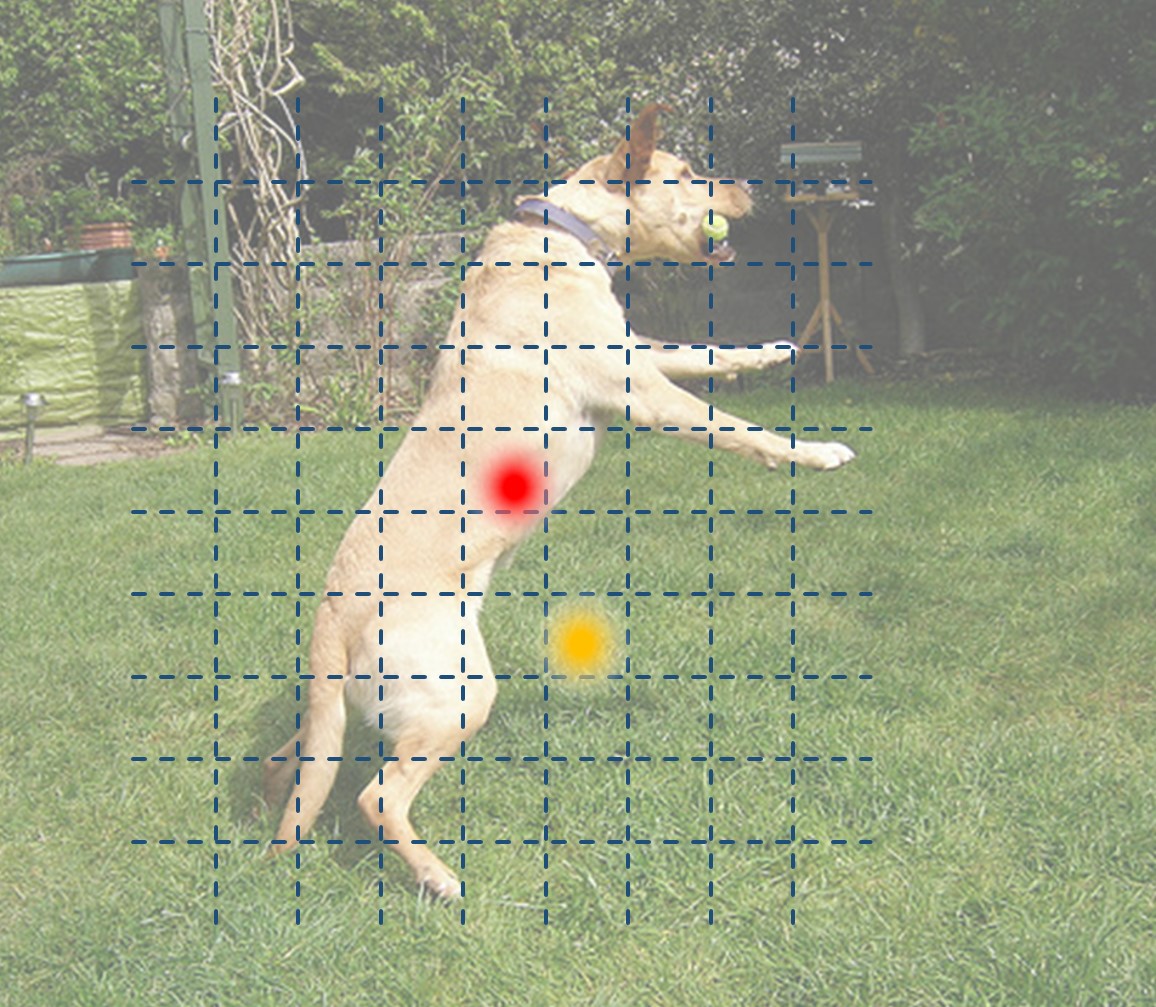}
    \end{minipage}
}
\caption{Demonstrative case of the feature misalignment between the original anchor and the refined anchor. (a) The green bounding box stands for the ground truth and the orange one represents the original anchor. The refined anchor is shown as the red bounding box. (b) Location of center points for original and refined anchors in the feature grid. Simply extracting features from the previous location (orange point) is inaccurate.}
\label{fig:misalignment}
\vspace{-0.3cm}
\end{figure}

In cascade manner, anchors are transformed to different positions after applying the regression offsets. As shown in Figure~\ref{fig:misalignment} (a), the original anchor (orange box) is regressed to the red one as the result. From the perspective of the feature grid (b), anchor feature is extracted from the orange point using a small fully convolutional network. If we simply add new stages based on the same feature map, it means that the feature of the refined anchor is still extracted from the orange point, leading to feature inconsistency. 
The misalignment of anchor feature and anchor position will severely harm the detection performance. To maintain the feature consistency between different stages, the features of the refined anchors should be adapted to new locations.  

\section{Cascade RetinaNet}

In this section, we first review the RetinaNet and then introduce the proposed Cas-RetinaNet, which is a unified network with cascaded heads attached to RetinaNet. The overall architecture is illustrated in Figure~\ref{fig:architecture}.

\subsection{RetinaNet}

RetinaNet~\cite{FocalLoss} is a representative architecture of single-stage detection approaches with state-of-the-art performance. It can be divided into the backbone network and two task-specific subnetworks. Feature Pyramid Network (FPN) is adopted as the backbone network for constructing a multi-scale feature pyramid efficiently. On top of the feature pyramid, classification subnet and box regression subnet are utilized for predicting categories and refining the anchor locations, respectively. Parameters of the two subnets are shared across all pyramid levels for efficiency. Due to the extreme foreground-background class imbalance, Focal Loss is adopted to prevent the vast number of easy negatives from overwhelming the detector during training.

\subsection{Cas-RetinaNet}

\begin{figure}[!t]
\centering
\includegraphics[width=0.75\linewidth]{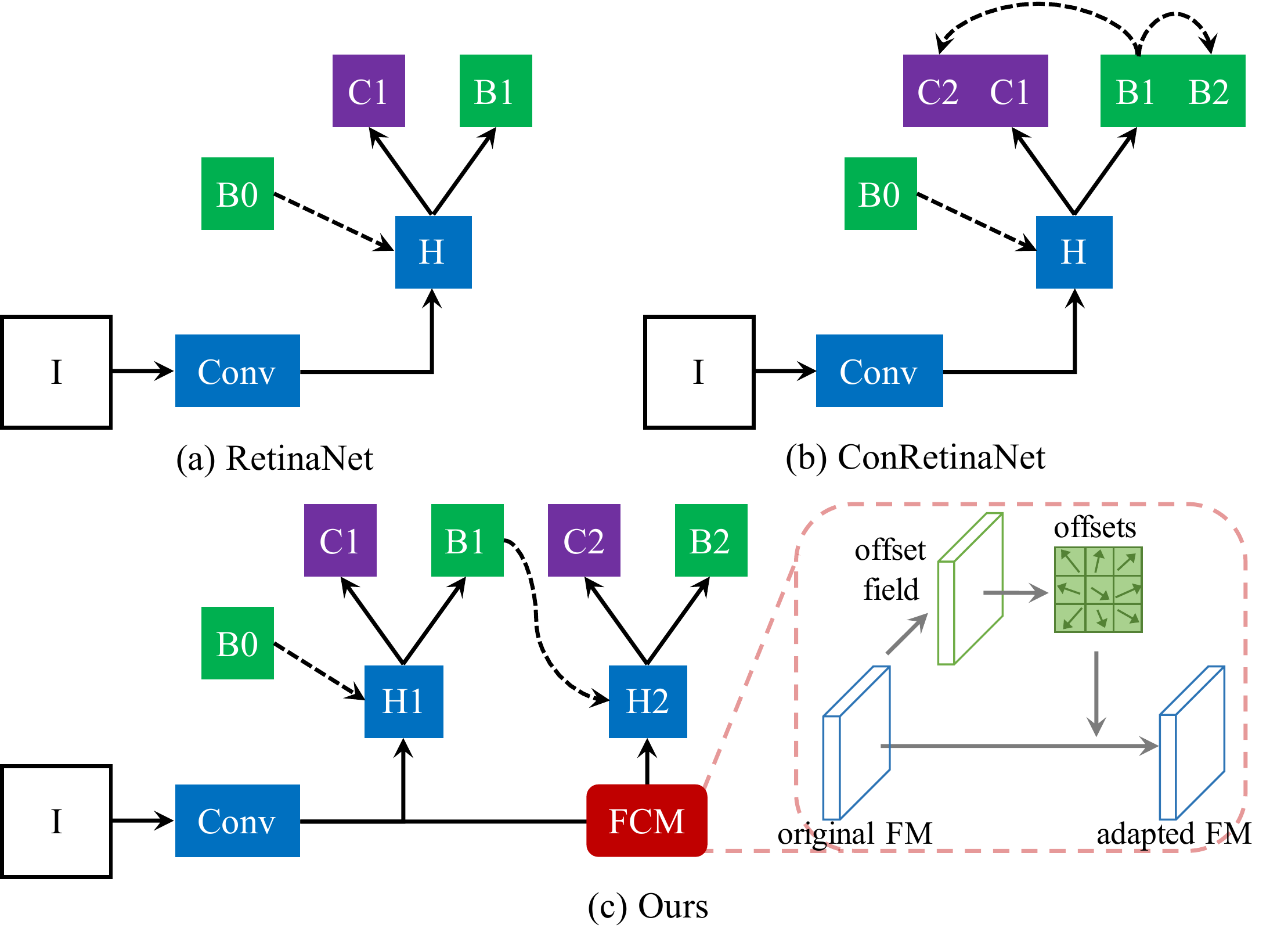}
\caption{Different architectures of single-stage detection frameworks. ``I'' is input image, ``conv'' backbone convolutions, ``H'' fully convolutional network head, ``B0'' pre-defined anchor box, ``C'' classification, ``B1, B2'' the refined anchor for different stages. Adapted feature map (``FM'') is generated using FCM for feature consistency.}
\label{fig:architecture}
\end{figure}

\textbf{Cascaded detection.} The difficult detection task can be decomposed into a sequence of simpler stages in a cascaded manner. Outputs from the previous stage are viewed as the input of the following stage. Generally speaking, the loss function for the $i$-th stage can be formulated as
\begin{equation}
    \mathcal{L}^i=\mathcal{L}_{cls}(c_i(x^i), y^i)+\lambda^i \mathbbm{1}[y^i\ge 1]\mathcal{L}_{loc}(r_i(x^i, b^i), g),
\end{equation}
where $x^i, c_i$ and $r_i$ stand for the backbone features, classification head and regression head for the $i$-th stage, respectively. $b^i$ and $g$ represent the predicated and ground truth bounding boxes, and $b^0$ the pre-defined anchors. Anchor labels $y^i$ are determined by calculating the IoU between $b^i$ and $g$. Specifically, $b^i$ are assigned to ground-truth object boxes using an IoU threshold of $T^i_{+}$; and to background if their IoU is in $[0, T^i_{-})$. As each input box is assigned to at most one object box, $y^i$ are obtained by turning the class label into the one-hot vector. Unassigned samples are ignored during the training process. Based on this, original Focal Loss and $\mathrm{Smooth}_{L1}$ loss~\cite{FocalLoss} are adopted as $\mathcal{L}_{cls}$ and $\mathcal{L}_{reg}$. The indicator function $\mathbbm{1}[y^i\ge 1]$ equals to 1 when $y^i\ge 1$ and 0 otherwise. $\lambda^i$ is the trade-off coefficient and is set to 1 by default. The overall loss function for cascade detection becomes
\begin{equation}
    \mathcal{L}=\alpha_1\mathcal{L}^1+\alpha_2\mathcal{L}^2+\cdots+\alpha_i\mathcal{L}^i+\cdots+\alpha_N\mathcal{L}^N.
\end{equation}
Trade-off coefficients $\alpha_1,\cdots,\alpha_N$ are set to 1 by default.

\textbf{Consistency between classification and localization.} As analyzed in Section~\ref{sec:misaligned_cls}, there is a huge gap between the classification confidence and localization performance in the first stage. The main reason lies in the sampling method as it decides the training examples as well as their weights. To be specific, $y^1$ are set to the class label if $IoU(b^1, g)\ge T^1_{+}(0.5)$ no matter the actual IoU is 0.55 or 0.95. A simple idea is gradually increasing the foreground IoU thresholds to constrain the classification confidence to be consistent with localization performance. We empirically increase the IoU threshold for the following stages such as $T_{+}^2=0.6$. As shown in Figure~\ref{fig:distribution} (b), feature consistency between classification and localization is improved. Note that the regression targets for $b^{i-1}$ and $b^i$ can be different, we re-assign the boxes to new ground truths at each different stage. Corresponding classification labels and regression targets are generated using the specified thresholds.

\textbf{Feature Consistency Module.} From the formulation above, we predict the classification scores and regression offsets based on the backbone feature $x^i$. Current cascade detectors usually adopt the same $x$ in multiple stages, which introduces feature misalignment as the location shifts are not considered. From our perspective, we hope to \textit{encode the current localization into the features of next stage}, just like transforming the location from the orange point to the red one in Figure~\ref{fig:misalignment}. We propose a novel FCM to adapt the feature to the latest location. As illustrated in the right part of Figure~\ref{fig:architecture} (c), a transformation offset from the original position to the refined one is learned based on $x^i$, and a deformable convolutional layer is utilized to produce the adapted feature $x^{i+1}$. FCM can be formulated as follows:
\begin{equation}
    x^{i+1} = FCM(x^i) = Deformable(x^i, offset(x^i)).
\end{equation}
Specifically, a $1\times 1$ convolution layer if adopted on top of $x^i$ for generating offsets for the $3\times 3$ bins in deformable convolution~\cite{Deformable}. Then a $3\times 3$ deformable convolution layer takes $x^i$ and the offsets to produce a new feature map $x^{i+1}$. It should be noted that Guided Anchoring~\cite{GuidedAnchor} also adopts deformable convolutions to align the features, but the main purpose is to improve the inconsistent representation caused by the predicted irregular anchor shapes. From the experiments, we prove that our proposed FCM can steadily improve the detection performance in different settings.



\section{Experiments}

\subsection{Experimental Setting}

\textbf{Dataset and evaluation metric.} Experimental results are presented on the bounding box detection track of the challenging MS COCO benchmark~\cite{COCO}. Following the common practice~\cite{FocalLoss}, we use the COCO \texttt{trainval35k} split (union of 80k images from \texttt{train} and a random 35k subset of images from the 40k image \texttt{val} split) for training and report the detection performance on the \texttt{minival} split (the remaining 5k images from \texttt{val}). The COCO-style Average Precision (AP) is chosen as the evaluation metric which averages AP across IoU thresholds from 0.5 to 0.95 with an interval of 0.05.

\textbf{Implementation Details.} We adopt RetineNet~\cite{FocalLoss} with ResNet-50~\cite{ResNet} model pre-trained on ImageNet~\cite{ImageNet} dataset as our baseline. All models are trained on the COCO \texttt{trainval35k} and tested on \texttt{minival} with image short size at 600 pixels unless noted. Original settings of RetinaNet such as hyper-parameters for anchors and Focal Loss are followed for fairly comparison. For the additional stages, we follow the original architecture of RetinaNet head, except for the changes in IoU thresholds and the proposed FCM. Classification loss and regression loss are found to be unbalanced in our experiments, so $\lambda$ is set to 2 for each stage. At inference time, regression offsets from different cascade stages are applied sequentially to the original anchors. Classification scores from different stages are averaged as the final score to achieve more robust results. We conduct ablation studies and analyze the impact of our proposed Cas-RetinaNet with various design choices.

\subsection{Ablation Study}

\begin{table}[!t]
\begin{center}
\begin{tabular}{c|c|c|c|ccccc}
\hline
Method & Scale & IoU & $\mathrm{AP}$ & $\mathrm{AP}_{50}$ & $\mathrm{AP}_{60}$ & $\mathrm{AP}_{70}$ & $\mathrm{AP}_{80}$ & $\mathrm{AP}_{90}$ \\
\hline
\hline
RetinaNet~\cite{FocalLoss} & 600 & - & 34.0 & 52.5 & - & - & - & -\\
RetinaNet~\cite{FocalLoss} & 800 & - & 35.4 & 53.9 & - & - & - & -\\
\hline
Cas-RetinaNet & 600 & 0.5 & 33.8 & 52.3 & 48.1 & 41.5 & 29.8 & 11.2\\
Cas-RetinaNet & 600 & 0.6 & 34.4 & 52.5 & 48.5 & 41.9 & 30.5 & 11.7\\
Cas-RetinaNet & 600 & 0.7 & 34.4 & 52.0 & 48.1 & 41.7 & 31.3 & \textbf{12.5}\\
\hline
Cas-RetinaNet & 800 & 0.5 & 35.4 & 54.6 & 50.4 & 43.0 & 31.4 & 11.8\\
Cas-RetinaNet & 800 & 0.6 & \textbf{36.1} & \textbf{55.0} & \textbf{50.8} & \textbf{43.9} & \textbf{32.5} & \textbf{12.5}\\
\hline
\end{tabular}
\end{center}

\caption{Ablation study for different IoU thresholds on COCO \texttt{minival} set. ``IoU'' means the foreground IoU threshold for the second stage. ``AP'' stands for the primary challenge metric for COCO dataset. ``Scale'' means the short side of input images.}
\label{tab:threshold}
\end{table}

\textbf{Comparison with Different IoU Thresholds.}
Detection performances are compared under different IoU thresholds on COCO dataset in Table~\ref{tab:threshold}. We first prove that simply adding a new stage with the same setting brings no gains for the detection accuracy. AP drops slightly or keeps unchanged for the Cas-RetinaNet with IoU threshold 0.5. We argue that the reason mainly lies in the misaligned classifications like the distribution shown in Figure~\ref{fig:distribution} (a), due to the unchanged sampling method. When the foreground threshold is increased to 0.6 for the second stage, we observe a reasonable improvement ($33.8\rightarrow 34.4$). Here we also try a higher IoU threshold 0.7 for the second stage. It clearly shows that improvements focus on higher IoU thresholds such as $\mathrm{AP}_{90}$, while the $\mathrm{AP}_{50}$ drops slightly. From our perspective, higher foreground IoU threshold brings training samples with higher quality, while the quantity becomes fewer. For simplicity and robustness, We choose 0.6 as the foreground IoU threshold for the second stage. Further experiments with a different input scale indicate a similar conclusion and show the effectiveness of our method.

\begin{table}[!t]
\begin{center}
\begin{tabular}{c|c|c|c|ccccc}
\hline
Backbone & Scale & FCM & $\mathrm{AP}$ & $\mathrm{AP}_{50}$ & $\mathrm{AP}_{60}$ & $\mathrm{AP}_{70}$ & $\mathrm{AP}_{80}$ & $\mathrm{AP}_{90}$\\
\hline
\hline
ResNet-50 & 600 & & 34.4 & 52.5 & 48.5 & 41.9 & 30.5 & 11.7\\
ResNet-50 & 600 & \checkmark & 35.5 & 54.0 & 49.7 & 43.3 & 32.0 & 12.6\\
\hline
ResNet-50 & 800 & & 36.1 & 55.0 & 50.8 & 43.9 & 32.5 & 12.5\\
ResNet-50 & 800 & \checkmark & 37.1 & 56.3 & 52.2 & 45.3 & 33.5 & 12.8\\
\hline
ResNet-101 & 800 & & 37.9 & 56.8 & 52.8 & 46.0 & 34.9 & 13.9\\
ResNet-101 & 800 & \checkmark & \textbf{38.9} & \textbf{58.1} & \textbf{53.9} & \textbf{47.1} & \textbf{36.2} & \textbf{14.3}\\
\hline
\end{tabular}
\end{center}

\caption{Ablation study for FCM on COCO \texttt{minival} set. Settings can be referred as Table~\ref{tab:threshold}. Foreground IoU threshold is set to 0.6 for all experiments.}
\label{tab:FCM}
\end{table}

\textbf{Feature Consistency Module.}
We adopt various experiments under different backbone capacities and input scales to validate the effectiveness of our proposed FCM in Table~\ref{tab:FCM}. Misalignments are ubiquitous in cascaded single-stage detectors and limit the detection performance. Benefit from the adapted feature map produced by FCM, the performances under different settings are improved by $\sim$ 1 point steadily. Note that the deformable part in FCM requires longer time to converge, we extend training time to 1.5$\times$. It is a fair comparison since little improvements are observed for RetinaNet when training with a 2$\times$ setting~\footnote{https://github.com/facebookresearch/detectron}. From the experiments, we show that our proposed FCM is simple but effective since it only consists of a convolution for producing offsets and a convolution for capturing the effective features considering the misalignments.

\begin{table}[!t]
\begin{center}
\begin{tabular}{c|c|c|ccccc}
\hline
\#Stages & Test stage & $\mathrm{AP}$ & $\mathrm{AP}_{50}$ & $\mathrm{AP}_{60}$ & $\mathrm{AP}_{70}$ & $\mathrm{AP}_{80}$ & $\mathrm{AP}_{90}$\\
\hline
\hline
1 & 1 & 34.0 & 52.5 & - & - & - & -\\
2 & $\overline{1\sim 2}$ & \textbf{35.5} & \textbf{54.0} & \textbf{49.7} & \textbf{43.3} & \textbf{32.0} & 12.6\\
3 & $\overline{1\sim 2}$ & 35.0 & 53.1 & 49.1 & 42.5 & \textbf{32.0} & 12.6\\
3 & $\overline{1\sim 3}$ & 34.9 & 52.9 & 49.0 & 42.4 & 31.9 & \textbf{12.7}\\
\hline
\end{tabular}
\end{center}
\caption{Ablation study for number of stages on COCO \texttt{minival} set. $\overline{1\sim 3}$ indicates the ensemble result, which is the averaged score of the three classifiers with the 3rd stage boxes.}
\label{tab:numbers}
\end{table}

\textbf{Number of stages.}
The impact of the number of stages is summarized in Table~\ref{tab:numbers}. Adding a second detection stage improves the baseline detector by 1.5 points in AP. However, the addition of the third stage ($T^3_{+}=0.7$) leads to a slight drop in the overall performance, while it reaches the best performance for high IoU levels. Cascading two stages achieves the best trade-off for Cas-RetinaNet.

\textbf{Complexity and speed.}
The computational complexity of Cas-RetinaNet increases with the number of cascade stages. For each new stage, the additional complexity comes from both the FCM and the head part. Compared to the backbone, the increased computational cost is really small. We evaluate the inference speed for both original RetinaNet and Cas-RetinaNet with ResNet-50 on a single RTX 2080TI GPU. As for the majority setting (adding one new stage with image short size at 800 pixels), Cas-RetinaNet achieves about 10 FPS and the original RetinaNet is about 12.5 FPS. Note that we apply the same head part as RetinaNet for the new stages to highlight the inconsistency problem, we believe that the complexity can be reduced by simplifying the head design.

\subsection{Comparison to State-of-the-Art}

The proposed Cas-RetinaNet is compared to state-of-the-art object detectors (both one-stage and two-stage) in Table~\ref{tab:overall}. Standard COCO metrics are reported on the \texttt{test-dev} set. Cas-RetinaNet improves detection performance on RetinaNet consistently by $1.5\sim 2$ points, independently of the backbone. Under ResNet-101 backbone, our model achieves state-of-the-art performances and outperforms all other models without any bells or whistles.

\begin{table}[!t]
\begin{center}
\setlength{\tabcolsep}{3pt}
\begin{tabular}{l|c|ccc|ccc}
\hline
Method & Backbone & $\mathrm{AP}$ & $\mathrm{AP}_{50}$ & $\mathrm{AP}_{75}$ & $\mathrm{AP_S}$ & $\mathrm{AP_M}$ & $\mathrm{AP_L}$\\
\hline
\hline
\textit{Two-stage methods} & & & & & &\\
Faster R-CNN+++~\cite{ResNet}* & ResNet-101 & 34.9 & 55.7 & 37.4 & 15.6 & 38.7 & 50.9\\
Faster R-CNN by G-RMI~\cite{Speed-Accuracy-Trade-off} & Inception-ResNet-v2 & 34.7 & 55.5 & 36.7 & 13.5 & 38.1 & 52.0\\
Faster R-CNN w FPN~\cite{FPN} & ResNet-101 & 36.2 & 59.1 & 39.0 & 18.2 & 39.0 & 48.2\\
Faster R-CNN w TDM~\cite{TDM} & Inception-ResNet-v2 & 36.8 & 57.7 & 39.2 & 16.2 & 39.8 & 52.1\\
Mask R-CNN~\cite{MaskRCNN} & ResNet-101 & 38.2 & 60.3 & 41.7 & 20.1 & 41.1 & 50.2\\
Relation~\cite{RelationNet} & DCN-101 & 39.0 & 58.6 & 42.9 & - & - & -\\
\hline
\textit{One-stage methods} & & & & & &\\
YOLOv2~\cite{YOLO9000} & DarkNet-19 & 21.6 & 44.0 & 19.2 & 5.0 & 22.4 & 35.5\\
SSD513~\cite{SSD} & ResNet-101 & 31.2 & 50.4 & 33.3 & 10.2 & 34.5 & 49.8\\
YOLOv3~\cite{YOLOv3} & Darknet-53 & 33.0 & 57.9 & 34.4 & 18.3 & 35.4 & 41.9\\
DSSD513~\cite{DSSD} & ResNet-101 & 33.2 & 53.3 & 35.2 & 13.0 & 35.4 & 51.1\\
RetinaNet~\cite{FocalLoss} & ResNet-50 & 35.7 & 55.0 & 38.5 & 18.9 & 38.9 & 46.3\\
RefineDet512~\cite{RefineDet} & ResNet-101 & 36.4 & 57.5 & 39.5 & 16.6 & 39.9 & 51.4\\
GA-RetinaNet~\cite{GuidedAnchor} & ResNet-50 & 37.1 & 56.9 & 40.0 & 20.1 & 40.1 & 48.0\\
RetinaNet~\cite{FocalLoss} & ResNet-101 & 37.8 & 57.5 & 40.8 & 20.2 & 41.1 & 49.2\\
RetinaNet~\cite{FocalLoss}$\dagger$ & ResNet-101 & 39.1 & 59.1 & 42.3 & 21.8 & 42.7 & 50.2\\
ConRetinaNet~\cite{ConsistentOptimization}$\dagger$ & ResNet-101 & 40.1 & 59.6 & 43.5 & 23.4 & 44.2 & 53.3\\
CornerNet511~\cite{CornerNet} & Hourglass-104 & 40.5 & 56.5 & 43.1 & 19.4 & 42.7 & \textbf{53.9}\\
\hline
\textit{Ours} & & & & & &\\
Cas-RetinaNet & ResNet-50 & 37.4 & 56.6 & 40.7 & 20.9 & 40.3 & 47.5\\
Cas-RetinaNet & ResNet-101 & 39.3 & 59.0 & 42.8 & 22.4 & 42.6 & 50.0\\
Cas-RetinaNet$\dagger$ & ResNet-101 & \textbf{41.1} & \textbf{60.7} & \textbf{45.0} & \textbf{23.7} & \textbf{44.4} & 52.9\\
\hline
\end{tabular}
\end{center}
\caption{Cas-RetinaNet vs. other state-of-the-art two-stage or one-stage detectors (single-model and single-scale results). We show the results of our Cas-RetinaNet models based on Resnet-50 and Resnet-101 with 800 input size. ``$\dagger$'' indicates that model is trained with scale jitter and for 1.5$\times$ longer than original ones. The entries denoted by ``*'' used bells and whistles at inference.}
\label{tab:overall}
\end{table}

\subsection{Discussion}
An interesting question is how to compare cascade single-stage detectors with two-stage ones. Generally speaking, the main difference lies in whether using the region-crop layer. Region features are powerful but add a lot of complexity as the number of region of interest (RoIs) increases. In other words, cascade single-stage methods are more concise and flexible due to the fully convolutional architecture. As for feature extraction, the deformable convolution in Cas-RetinaNet aggregates features from other semantic points to generate ``region features''. Consequently, it will be a better framework for object detection.

\section{Conclusion}

In this paper, we take a thorough analysis of the single-stage detectors and point out two main designing rules for the cascade manner which lies in maintaining the consistency. A multistage object detector named Cas-RetinaNet is proposed to address these problems. Sequential stages trained with increasing IoU thresholds and a novel Feature Consistency Module are adopted to improve the inconsistency. We conduct sufficient experiments and the stable detection improvements on the challenging COCO dataset prove the effectiveness of our method. We believe that this work can benefit future object detection researches. 

\section*{Acknowledgement}
This work is partially supported by National Key R\&D Program of China (No.2017YFA0700800), Natural Science Foundation of China (NSFC): 61876171 and 61572465.

\bibliography{egbib}
\end{document}